\documentclass[10pt,hidelinks,conference]{IEEEtran}

\ifCLASSINFOpdf
\else
\usepackage[dvips]{graphicx}
\fi
\usepackage{url}

\usepackage{graphicx} 
\usepackage{booktabs} 
\usepackage{mathrsfs}
\usepackage{amsmath}
\usepackage{amssymb}
\usepackage{bm}
\usepackage{mathtools}
\usepackage{cite}
\usepackage[acronym,shortcuts]{glossaries}
\usepackage{enumitem} 
\usepackage{comment}
\usepackage{algorithm, algpseudocode}
\usepackage{mathtools}
\usepackage{lipsum}
\usepackage{mleftright}
\usepackage{orcidlink}

\setlist[itemize]{noitemsep} 

\DeclareMathOperator*{\argmin}{arg\,min}

\providecommand{\UL}{\mathrel{\raise-4pt\hbox{\hglue -2.8ex
\vrule height .1ex width 2.3ex
\vrule height 3ex width .1ex
\hglue .4ex}}}
\providecommand{\ul}{\mathrel{\raise-2pt\hbox{\hglue -2.3ex
\vrule height .1ex width 2ex
\vrule height 2ex width .1ex
\hglue .4ex}}}
\providecommand{\ub}{
\mathrel{\hbox{\hglue -2.8ex \vrule height 2ex width .06ex}
\raise2ex\hbox{\hglue -0.1ex\vrule height .1ex width 2.5ex}
\hbox{\hglue -0.1ex \vrule height 2ex width .1ex
\hglue .4ex}}}
\providecommand{\lb}{
\mathrel{\raise-0.5ex
\hbox{\hglue -2.8ex \vrule height 2ex width .1ex
\vrule height .1ex width 2.5ex
\vrule height 2ex width .1ex
\hglue .4ex}}}
\providecommand{\UB}{
\mathrel{
\raise-1ex\hbox{\hglue -1.8em \vrule height 3ex width .1ex}
\raise2ex\hbox{\hglue -0.1ex\vrule height .1ex width 1.5em}
\raise-1ex\hbox{\hglue -0.1ex \vrule height 3ex width .1ex
\hglue .4ex}}}
\providecommand{\LB}{
\mathrel{\raise-1ex
\hbox{\hglue -1.8em \vrule height 3ex width .1ex
\vrule height .1ex width 1.5em
\vrule height 3ex width .1ex
\hglue .4ex}}}
\usepackage{trimclip}
\newif\iflclip
\newif\ifbclip
\newif\ifrclip
\newif\iftclip
\def\CLIP{\dimexpr\fboxrule+.2pt\relax}
\def\nulclip{0pt}
\newcommand\partbox[2]{%
\lclipfalse\bclipfalse\rclipfalse\tclipfalse%
\let\lkern\relax\let\rkern\relax%
\let\lclip\nulclip\let\bclip\nulclip\let\rclip\nulclip\let\tclip\nulclip%
\parseclip#1\relax\relax%
\iflclip\def\lkern{\kern\CLIP}\def\lclip{\CLIP}\fi
\ifbclip\def\bclip{\CLIP}\fi
\ifrclip\def\rkern{\kern\CLIP}\def\rclip{\CLIP}\fi
\iftclip\def\tclip{\CLIP}\fi
\lkern\clipbox{\lclip{} \bclip{} \rclip{} \tclip}{\fbox{#2}}\rkern%
}
\def\parseclip#1#2\relax{%
\ifx l#1\lcliptrue\else
\ifx b#1\bcliptrue\else
\ifx r#1\rcliptrue\else
\ifx t#1\tcliptrue\else
\fi\fi\fi\fi
\ifx\relax#2\relax\else\parseclip#2\relax\fi
}
\usepackage{efbox}

\newcommand\Tstrut{\rule{0pt}{2.6ex}}         

\newacronym{SVD}{SVD}{singular value decomposition}
\newacronym{DCM}{DCM}{double-centering matrix}
\newacronym{3D}{3D}{three-dimensional}
\newacronym{GA}{GA}{genie-aided}
\newacronym{EA}{EA}{``\emph{estimate-then-average}''}
\newacronym{AE}{AE}{``\emph{average-then-estimate}''}
\newacronym{IRS}{IRS}{intelligent reflecting surface}
\newacronym{RSSI}{RSSI}{received signal strength indicator}
\newacronym{SotA}{SotA}{state-of-the-art}
\newacronym{CSI}{CSI}{channel state information}
\newacronym{D2D}{D2D}{device-to-device}
\newacronym{RR}{RR}{round-robin}
\newacronym{DA}{DA}{Dutch auction}
\newacronym{AV}{AV}{autonomous vehicle}
\newacronym{CWFL}{CWFL}{clustered WFL}
\newacronym{WFL}{WFL}{wireless federated learning}
\newacronym{RSMA}{RSMA}{rate splitting multiple access}
\newacronym{IoT}{IoT}{Internet-of-Things}
\newacronym{TDMA}{TDMA}{time-domain multiple access}
\newacronym{NOMA}{NOMA}{non-orthogonal multiple access}
\newacronym{ML}{ML}{machine learning}
\newacronym{MIMO}{MIMO}{multiple-input multiple-output}
\newacronym{CT}{CT}{compute-then-transmit}
\newacronym{FP}{FP}{fractional programming}
\newacronym{CF-mMIMO}{CF-mMIMO}{cell free massive MIMO}
\newacronym{iid}{i.i.d.}{independent and identically distributed}
\newacronym{AD}{AD}{autonomous driving}
\newacronym{DL}{DL}{downlink}
\newacronym{UL}{UL}{uplink}
\newacronym{IC}{IC}{interference cancellation}
\newacronym{SIC}{SIC}{successive interference cancellation}
\newacronym{BS}{BS}{base station}
\newacronym{TX}{TX}{transmit}
\newacronym{RX}{RX}{receive}
\newacronym{MU}{MU}{multi-user}
\newacronym{SISO}{SISO}{single-input single-output}
\newacronym{AWGN}{AWGN}{additive white Gaussian noise}
\newacronym{SINR}{SINR}{signal-to-interference-and-noise ratio}
\newacronym{FL}{FL}{federated learning}
\newacronym{CPU}{CPU}{central processing unit}
\newacronym{KNN}{KNN}{K-nearest-neighbor}
\newacronym{RF}{RF}{radio frequency}
\newacronym{GD}{GD}{gradient descent}
\newacronym{V2X}{V2X}{vehicle-to-anything}

\newacronym{RSS}{RSS}{received signal strength}
\newacronym{FIM}{FIM}{fisher information matrix}
\newacronym{ToA}{ToA}{time of arrival}
\newacronym{ToF}{ToF}{time of flightl}
\newacronym{AoA}{AoA}{angle of arrival}
\newacronym{GP}{GP}{Gaussian process}
\newacronym{2D}{2D}{two-dimensional}
\newacronym{GPR}{GPR}{Gaussian process regression}
\newacronym{GNSS}{GNSS}{global navigation satellite systems}
\newacronym{B5G}{B5G}{beyond fifth-generation}
\newacronym{RRH}{RRH}{remote radio head}
\newacronym{GPS}{GPS}{Global Positioning System}
\newacronym{RFID}{RFID}{radio frequency identification}
\newacronym{TCAS}{TCAS}{traffic alert and collision avoidance systems}
\newacronym{RMSE}{RMSE}{root mean square error}
\newacronym{SGD}{SGD}{stochastic gradient descent}
\newacronym{PDF}{PDF}{probability density function}
\newacronym{CU}{CU}{computing unit}
\newacronym{DM-MIMO}{DM-MIMO}{distributed massive multiple-input multiple-output}
\newacronym{LOS}{LOS}{line-of-sight}
\newacronym{NLOS}{NLOS}{non-line-of-sight}
\newacronym{ROI}{ROI}{region of interest}
\newacronym{AP}{AP}{access point}
\newacronym{TDOA}{TDOA}{time difference of arrival}
\newacronym{UE}{UE}{user equipment}
\newacronym{dB}{dB}{decibel}
\newacronym{RIS}{RIS}{reconfigurable intelligent surface}
\newacronym{CG}{CG}{conjugate gradient}

\newacronym{PG}{PG}{proximal gradient}
\newacronym{SVT}{SVT}{singular value thresholding}
\newacronym{NN}{NN}{nuclear norm}
\newacronym{NMSE}{NMSE}{normalized mean square error}
\newacronym{MC}{MC}{matrix completion}
\newacronym{NP}{NP}{non-deterministic polynomial-time}
\newacronym{EDM}{EDM}{euclidean distance matrix}
\newacronym{SC}{SC}{soft-connected}
\newacronym{CRLB}{CRLB}{Cramér-Rao Lower Bound}
\newacronym{PoA}{PoA}{phase of arrival}
\newacronym{UAV}{UAV}{unmanned aerial vehicle}
\newacronym{VR}{VR}{virtual reality}
\newacronym{MDS}{MDS}{multidimensional scaling}

\newacronym{RBL}{RBL}{rigid body localization}
\newacronym{RBT}{RBT}{rigid body tracking}
\newacronym{SC-RBL}{SC-RBL}{soft-connected RBL}
\newacronym{W-RBL}{W-RBL}{\underline{wireless} RBL}

\newacronym{SDP}{SDP}{semidefinite programming}
\newacronym{JCAS}{JCAS}{joint communication and sensing}
\newacronym{SDR}{SDR}{semi-definite relaxation}

\newacronym{OPP}{OPP}{orthogonal Procrustes problem}
\newacronym{SLAM}{SLAM}{simultaneous localization and mapping}
\newacronym{WLS}{WLS}{weighted least square}
\newacronym{SI}{SI}{soft-impute}

\begin{document}

\title{Egoistic MDS-based Rigid Body Localization\vspace{-.2ex}}

\author{\IEEEauthorblockN{Niclas~F\"uhrling*\textsuperscript{\orcidlink{0000-0003-1942-8691}}, Giuseppe Abreu*\textsuperscript{\orcidlink{0000-0002-5018-8174}}, David~Gonz{\'a}lez~G.$^\dag$\textsuperscript{\orcidlink{0000-0003-2090-8481}} and Osvaldo~Gonsa$^\dag$\textsuperscript{\orcidlink{0000-0001-5452-8159}}}
\IEEEauthorblockA{\textit{*School of Computer Science and Engineering, Constructor University, Bremen, Germany} \\ \textit{$^\dag$Wireless Communications Technologies Group, Continental AG, Frankfurt, Germany} \\ 
(nfuehrling,gabreu)@constructor.university, david.gonzalez.g@ieee.org, osvaldo.gonsa@continental-corporation.com\\[-3ex]}
}

\setlength{\parskip}{0pt}

\maketitle

\begin{abstract}
We consider a novel anchorless \ac{RBL} suitable for application in \ac{AD}, in so far as the algorithm enables a rigid body to egoistically detect the location (relative translation) and orientation (relative rotation) of another body, without knowledge of the shape of the latter, based only on a set of measurements of the distances between sensors of one vehicle to the other.
A key point of the proposed method is that the translation vector between the two-bodies is modeled using the double-centering operator from \ac{MDS} theory, enabling the method to be used between rigid bodies regardless of their shapes, in contrast to conventional approaches which require both bodies to have the same shape.
Simulation results illustrate the good performance of the proposed technique in terms of \ac{RMSE} of the estimates in different setups.
\end{abstract}


\IEEEpeerreviewmaketitle

\vspace{-1ex}
\section{Introduction}

\IEEEPARstart{W}{ireless} localization \cite{Yassin_2016} can be seen as a precursor of \ac{JCAS}, demonstrating how communication signals can also be used for sensing an environment, including localization of users.
There are many types of information that can be extracted from radio signals for the purpose of localization, including finger-prints \cite{VoCST2016}, \ac{RSSI} \cite{Nic:RSSI}, \ac{AoA} \cite{Al-SadoonTAP2020}, or delay-based estimates of radio range \cite{ZengTSP2022}.
Conventionally, such information needed for localization was generally assumed to be obtained by specialized equipment and protocols, requiring the transmission of dedicated signals, implicating in costs and other constraints which in turn explains the predominance in related literature \cite{Yassin_2016,burghal_2020} of methods to find the position of individual points.  

Recently, however, advances in \ac{JCAS} technology \cite{Zhang_2021} has demonstrated that radar parameters ($i.e.$, range, bearing and velocity) can be acquired by conventional communications signals \cite{Rayan_2024,Rayan_Journal}, not only actively, $i.e.$, using signals transmitter by the target to the sensors, but also passively,  $i.e.$, using round-trip reflections of signals transmitted by the sensors themselves, which in turn implies a more abundant and richer availability of positioning information.
A consequence of this development is an increasing interest in the \acf{RBL} problem \cite{WangTSP2020,FuehrlingV2X2024}, whose objective is to determine not only the average location of targets, but their shape and orientation, based
on a collection of points sufficient to define the object. 
This feature of \acf{RBL} is particularly attractive to \ac{V2X} networks, where --
unlike earlier applications of positioning technology such as asset management in industrial settings \cite{AHMED_2020} and people tracking in indoor settings -- information on the size, shape, and orientation of vehicles are crucial to ensure the efficacy and safety of \acf{AD} applications such as collision detection \cite{Bruk_2023}, navigation \cite{eckenhoff_2019}, and vehicle path prediction \cite{Huang_2022}, to name only a few examples.

It is important also to distinguish between the type of \ac{RBL} system here addressed, which is based on radio signals, possibly under a \ac{JCAS} paradigm \cite{Chen_2015,WangTSP2020}, and conventional \ac{SLAM} technologies \cite{Huang_2019,Barros_2022,Bavle_2023} relying on dedicated equipment and  massive amounts of data \cite{Seref_2013} to function, which makes the latter less likely to be useful in the day-to-day \ac{AD} applications envisioned for a future where \acp{AV} will be widely deployed.

%
%
An example of the radio-based \ac{RBL} approach which is the subject of this article is the method in \cite{Bras_2016}, where the pose, angular velocity and trajectory of a rigid body is estimated using Lyapunov functions of Doppler measurements, obtained by a nonlinear observer.
Another example is \cite{Chen_2015}, in which a two-stage approach was used to estimate rotation, translation, angular velocity and translational velocity by range and Doppler measurements, making use of various \ac{WLS} minimization methods.
And going beyond the problem of \ac{RBL} involving a single object, the scheme in \cite{PizzoICASSP2016} which proposes a new relative multi-object \ac{RBL} method\footnote{An anchor-based version of the method had been proposed earlier in \cite{Chepuri_2013}.}
 in an anchorless scenario, whereby the relative translation and rotation between two rigid bodies is estimated by measuring the cross-body \ac{LOS} distances between the points defining the two bodies.

The latter case relates to a common scenario in \ac{AD} where a vehicle is able to measure the distance between itself and vehicles in its surroundings, such that the corresponding \ac{RBL} solution would find a large a direct and crucial application.
Unfortunately, however, most \ac{SotA} \ac{RBL} methods assume that the shape of the target rigid body is known \cite{Chepuri_2013, Chen_2015, PizzoICASSP2016}, which is unrealistic in real life applications since vehicles vary greatly in shape and size.

In view of the above, we propose in this article an anchorless and \ac{MDS}-based egoistic approach for \ac{RBL}, in which a rigid body ($e.g.$ vehicle) can estimate not only the distance, but also the shape and orientation of another ($e.g.$ another vehicle, possibly of different size and shape), based only on cross-body sensor-to-sensor range measurements.

The structure of the remainder of article is as follows.
First, a description of the system and measurement model is offered in Section \ref{sec:prior}.
Then, in Section \ref{sec:prop}, the proposed method including the convex optimization problem for the estimation of the translation, shape and orientation of the target rigid body is introduced, after a brief introduction of the \ac{SotA} \ac{MDS}-based \ac{RBL} approach.
Finally, a comparison of the proposed scheme with the conventional least square-based \ac{RBL} technique of \cite{Chen_2015}, adjusted to an egoistic setting, is offered in Section \ref{sec:res}, followed by further performance evaluation in terms of \ac{RMSE} results with different parameters.

\vspace{-1ex}
\section{Rigid Body Localization System Model}
\label{sec:prior}

\subsection{System Model}

\begin{figure}[H]
\centering
\includegraphics[width=\columnwidth]{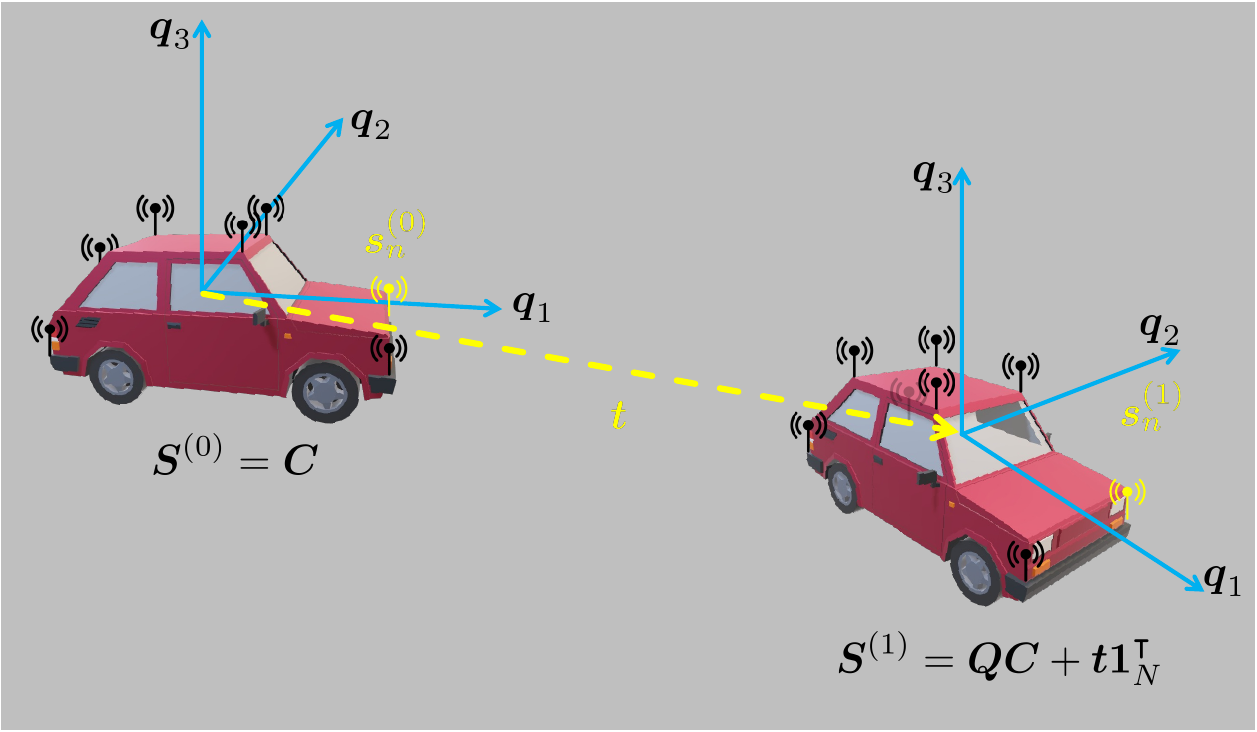}
\vspace{-4ex}
\caption{Illustration of a rigid body at two distinct locations $\boldsymbol{S}^{(0)}$ and $\boldsymbol{S}^{(1)}$. Without loss of generality, we set the initial to be identical to the matrix $\boldsymbol{C}$, which defines the shape and orientation of the rigid body. The second location $\boldsymbol{S}^{(1)}$ of the body relative to its initial location $\boldsymbol{S}^{(0)}$ is then determined according to equation \eqref{eq:basic_model_one_body}, and is obtained by the transformation of $\boldsymbol{S}^{(0)}$ via a rotation matrix $\bm{Q}$ and a translation vector $\boldsymbol{t}$.}
\label{fig:RB_tra}
\vspace{1ex}
\includegraphics[width=\columnwidth]{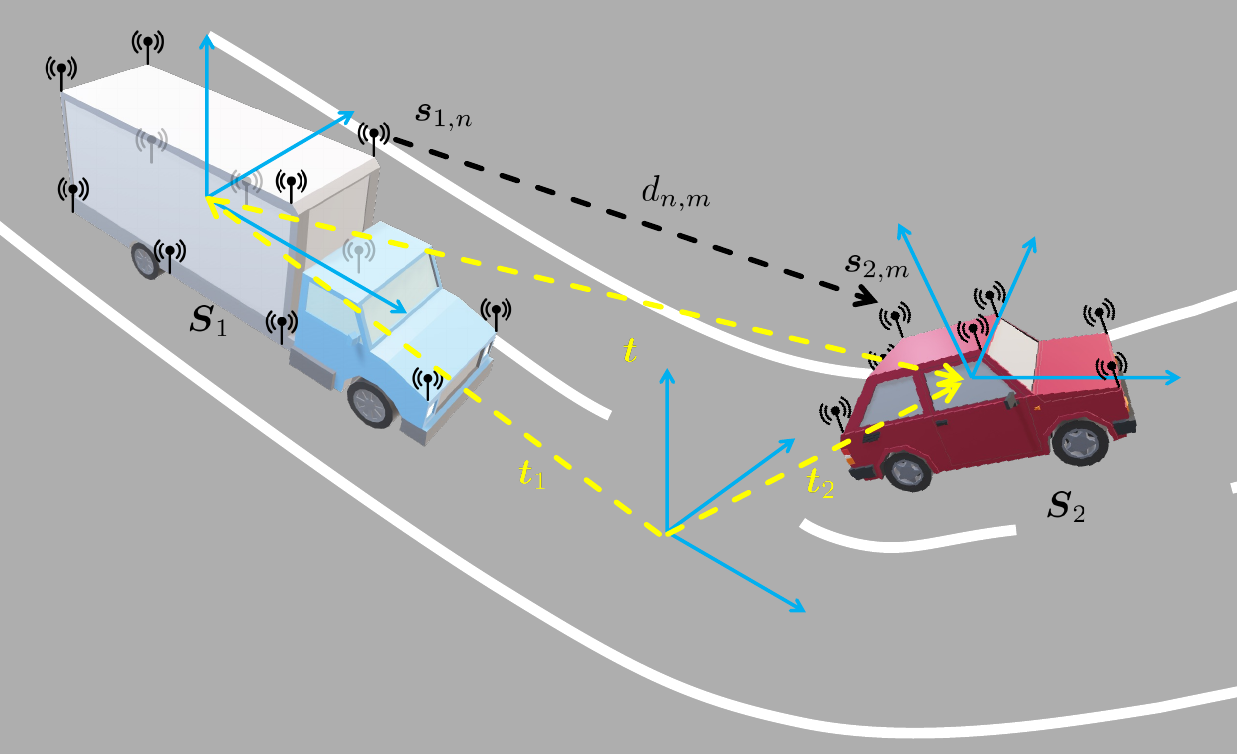}
\vspace{-4ex}
\caption{Illustration of two-body egoistic \ac{RBL} scenario. Each rigid body has a different shape, defined by distinct conformation matrices $\boldsymbol{C}_1$ and $\boldsymbol{C}_2$, respectively. The translation vector $\boldsymbol{t}$ between the bodies, depicted in yellow, is defined by the difference between the geometric centers of the two bodies.}
\label{fig:sys}
\vspace{-2ex}
\end{figure}

Referring to the illustration in Figure \ref{fig:RB_tra}, let a given rigid body be represented by a collection of $N$ landmark points $\boldsymbol{c}_n\in\mathbb{R}^{3\times 1}$ in the \ac{3D} space, with $n=\{1,\cdots,N\}$, such that the shape of said body is well described by the corresponding conformation matrix $\boldsymbol{C}$ constructed by the column-wise collection of the vectors $\boldsymbol{c}_n$.
Then, consider the representation of the location $\boldsymbol{S}^{(1)}$ of said rigid body relative to another location ($e.g.$, earlier location in case the body is in motion) $\boldsymbol{S}^{(0)}$, which without loss of generality can be set to be a ``canonical'' reference (centered at the absolute origin),
such that $\boldsymbol{S}^{(0)}=\boldsymbol{C}$ and thus one can write\footnote{Hereafter we drop super-scripts, which are no longer necessary.}
\vspace{-0.5ex}
\begin{equation}
\label{eq:basic_model_one_body}
\boldsymbol{S}=\boldsymbol{Q}\cdot\boldsymbol{C}+\boldsymbol{t}\cdot\boldsymbol{1}_{N}^{\intercal}=\mleft[\boldsymbol{Q}|\boldsymbol{t}\mright]\mleft[
\begin{array}{c}
\bm{C} \\
\hline
\boldsymbol{1}_{N}^{\intercal}
\end{array}
\mright],
\vspace{-0.5ex}
\end{equation}
where $\boldsymbol{t}\in\mathbb{R}^{3\times 1}$ is a translation vector given by the difference of the geometric centers of the body at the two locations, $\boldsymbol{1}_{N}$ is a column vector with $N$ entries all equal to $1$, and $\boldsymbol{Q}\in \mathbb{R}^{3\times 3}$ is a rotation matrix\footnote{For the sake of simplicity, in this article, detecting the orientation of a rigid body will be interpreted as estimating of the $9$ elements of the corresponding rotation matrix $\bm{Q}$ as a whole. In a follow up work, however, this will be extended by replacing the estimation of $\bm{Q}$ with the estimation of the associated and fundamental yaw, pitch and roll angles $(\alpha, \beta, \gamma)$.} determined by corresponding yaw, pitch and roll angles $\alpha, \beta$ and $\gamma$, respectively, namely
\vspace{-0.5ex}
\begin{eqnarray}
\bm{Q} \triangleq \bm{Q}_{z}(\gamma)\,\bm{Q}_{y}(\beta)\,\bm{Q}_{x}(\alpha)&&\\
&&\hspace{-28ex}
=\!\!\!\left[
\begin{array}{@{}c@{\;\,}c@{\;\,}c@{}}
\cos\gamma&-\sin\gamma& 0\\
\sin\gamma& \cos\gamma& 0\\
0 	    & 0           & 1\\
\end{array}\right]\!\!\!\cdot\!\!\!
\left[
\begin{array}{@{}c@{\;\,}c@{\;\,}c@{}}
\cos\beta & 0           & \sin\beta\\
0			& 1			  & 0\\
-\sin\beta& 0 		  & \cos\beta\\
\end{array}\right]\!\!\!\cdot\!\!\!
\left[
\begin{array}{@{\,}c@{\;\,}c@{\;\,}c@{\!}}
1 			& 0			  & 0\\
0			& \cos\alpha& -\sin\alpha\\
0			& \sin\alpha& \cos\alpha\\
\end{array}\right]\nonumber\\
&&\hspace{-28ex}
=\!\!\text{\scalebox{0.75}{$
\left[
\begin{array}{@{\,}c@{\;\,}c@{\;\,}c@{\,}}
\cos\beta\cos\gamma & \sin\alpha\sin\beta\cos\gamma- \cos\alpha\sin\gamma & \cos\alpha\sin\beta\cos\gamma+\sin\alpha\sin\gamma\\
\cos\beta\sin\gamma & \sin\alpha\sin\beta\sin\gamma+ \cos\alpha\cos\gamma & \cos\alpha\sin\beta\sin\gamma-\sin\alpha\cos\gamma\\
-\sin\beta			& \sin\alpha\cos\beta								  & \cos\alpha\cos\beta\\
\end{array}\right]$}}\nonumber\\
&&\hspace{-28ex}
=\!\!\left[
\begin{array}{@{\,}c@{\;\,}c@{\;\,}c@{\,}}
q_{1,1} & q_{1,2} & q_{1,3}\\
q_{2,1} & q_{2,2} & q_{2,3}\\
q_{3,1} & q_{3,2} & q_{3,3}\\
\end{array}\right]\!\!.\nonumber
\end{eqnarray}

Next, consider a scenario as illustrated in Figure \ref{fig:sys}, in which two rigid bodies, hereafter referred to by their indices $i=\{1,2\}$, have generally different shapes and/or are characterized by generally distinct numbers $N_1$ and $N_2$ of landmark points, respectively, such that under a common absolute reference, the bodies are represented by the corresponding distinct conformation matrices $\bm{C}_1\in\mathbb{R}^{3\times N_1}$ and $\bm{C}_2\in\mathbb{R}^{3\times N_2}$.

Since $\bm{C}_1 \neq \bm{C}_2$, it is obvious that in such a scenario the location of one body relative to the other cannot be described in terms of equation \eqref{eq:basic_model_one_body}.
A common problem in \ac{V2X} systems with relevance to \ac{AD} applications is, however, that one rigid body -- say, the truck in Figure \ref{fig:sys} -- is able to estimate not only its distance to the other -- in this case, the car in Figure \ref{fig:sys} -- but also its shape and orientation, based on a set of measurements of the distances between their corresponding landmark points.

It will be considered, in what follows, that such measurements can be obtained by deploying to the landmark points of each rigid body, a set of wireless transceivers, hereafter referred to as ``sensors'', capable of performing such mutual distance estimates\footnote{As discussed in the introduction, the setup with sensors deployed in each rigid body can be replaced by one in which each rigid body is equipped with radar or \ac{JCAS} technology capable of measuring distances from a set of points in one body to another set of points in the other. Mathematically, however, both approaches to the techniques utilized and proposed in this article.}.
It will, furthermore, be assumed that each body is only aware of its own shape, described by corresponding conformation matrices $\boldsymbol{C}_i=[\boldsymbol{c}_{i,1},\cdots,\boldsymbol{c}_{i,N_i}]\in \mathbb{R}^{3\times N_i}$, where $\boldsymbol{c}_{i,n}$, is the location of the $n$-th point of the $i$-th body, with respect to its geometric center.

\subsection{Measurement Model}

When subject to unbiased estimation errors, the estimates of the distance between a pair of sensors $\bm{s}_{1,n}$ on the first body, and $\bm{s}_{2,m}$ on the second\footnote{Without lack of clarity, we abuse the notation slightly by using $\bm{s}_{i,n}$ in reference both to a sensor and its location.}, can be described by
\vspace{-0.5ex}
\begin{equation}
\label{eq:distance_measurements}
\tilde{d}_{n,m} = d_{n,m} + \upsilon_{n,m},
\vspace{-0.5ex}
\end{equation}
where $d_{n,m} \triangleq ||\boldsymbol{s}_{1,n}-\boldsymbol{s}_{2,m}||_2$ is the true pairwise distance between the sensors, while $\upsilon_{n,m}$ denotes noise modeled as i.i.d. zero mean Gaussian random variables with variance $\sigma^2$.

In order to avoid negative numbers and linearize the relationship between the acquired squared distances and corresponding measurement errors, we shall also consider the equivalent model
\vspace{-1.5ex}
\begin{equation}
\label{eq:sqr_meas}
\tilde{d}^2_{n,m} = d^2_{n,m} + \omega_{n,m},
\vspace{-0.5ex}
\end{equation}
where the mean and variance of the measurement error $\omega_{n,m}$ are respectively given by $\mathbb{E}[\omega_{n,m}]=\sigma^2$ and $\mathbb{E}\big[\big(\omega_{n,m}-\mathbb{E}[\omega_{n,m}]\big)^2\big]=4d^2_{n,m}\sigma^2+2\sigma^4$, as described in \cite{Chepuri_2013}.

It proves convenient, to collect the true distances $d_{n,m}$ from above into the \ac{EDM}
\begin{equation}
\bm{D}=\mleft[
\begin{array}{c|c}
\bm{D}_1 & \bm{D}_{12} \\
\hline
\bm{D}_{12}^{\intercal}&\bm{D}_2
\end{array}
\mright] \in\mathbb{R}^{(N_1+N_2)\times (N_1+N_2)}.
\label{eq:full_D}
\end{equation}

\subsection{Problem Statement}

With the system and measurement model, we are ready to clearly define the problem we seek to solve and, for the sake of context, discuss a particularly relevant \ac{SotA} method.
To that end, let us first observe that assuming, without loss of generality, that the rigid body 1 ($i.e.$, the truck) attempts to egoistically locate body 2 ($i.e.$, the car), the system model conditions described earlier translate to the assumption that the self intra-distance matrix $\bm{D}_1$ is known exactly, the target intra-distance matrix $\bm{D}_2$ is unknown, and the squared cross-distance matrix $\bm{D}_{12}$ can be written as
\begin{equation}
\bm{D}_{12}^{\odot 2}=\bm{D}_{12}\odot \bm{D}_{12}=\boldsymbol{\psi}_1\boldsymbol{1}_{N_2}^{\intercal}+\boldsymbol{1}_{N_1}\boldsymbol{\psi}_2^{\intercal}-2\boldsymbol{S}_{1}^{\intercal}\boldsymbol{S}_{2},
\label{eq:meas_full}
\end{equation}
where $\boldsymbol{S}_{1}$ and $\boldsymbol{S}_{2}$ are matrices containing the locations of the sensors in bodies 1 and 2, respectively, the auxiliary vectors $\boldsymbol{\psi}_i \triangleq \boldsymbol{S}_{i}^{\intercal}\boldsymbol{S}_{i}$ carry the squared norms of the corresponding individual sensor locations, and the symbol $\odot$ indicates an element-wise matrix operation ($e.g.$, multiplication or exponentiation).

Next, consider an augmented sensor location matrix carrying the positions of all landmark points in both bodies, such that we may write, in similarity to equation \eqref{eq:basic_model_one_body} 
\begin{equation}
\label{eq:big_S}
\!\boldsymbol{S}\!=\![\boldsymbol{S}_{1}|\boldsymbol{S}_{2}]\!=
\![\boldsymbol{Q}_{1}|\boldsymbol{Q}_{2}]\!
\mleft[
\begin{array}{@{\,}c@{\,}|@{\,}c@{\,}}
\boldsymbol{C}_1 & \boldsymbol{0}_{3\times {N_2}} \\
\hline
\boldsymbol{0}_{3\times {N_1}}&\boldsymbol{C}_2
\end{array}
\mright]\!\! + \![\boldsymbol{t}_{1}|\boldsymbol{t}_{2}]\!\mleft[
\begin{array}{@{\,}c|c@{\,}}
\boldsymbol{1}_{N_1}^{\intercal} & \boldsymbol{0}_{N_2}^{\intercal} \\[0.5ex]
\hline
\boldsymbol{0}_{N_1}^{\intercal}&\boldsymbol{1}_{N_2}^{\intercal}
\end{array}
\mright]\!,\!\!\!\!
\end{equation}
where $\boldsymbol{Q}_{i}$ and $\boldsymbol{t}_{i}$ respectively denote the rotation matrix and translation vector of the $i$-th body, while $\boldsymbol{0}_{3\times N}$, $\boldsymbol{0}_{N}$ and $\boldsymbol{1}_N$ denote an all-zero matrix and an all-zero/all-one column vector, respectively.

Under the egoistic assumptions that $\boldsymbol{S}_{1}=\boldsymbol{C}_{1}$, and $\boldsymbol{t}_{1} = \boldsymbol{0}_3$, however, equation \eqref{eq:big_S} reduces to
\newpage
\begin{equation}
\label{eq:big_S_ego}
\boldsymbol{S}\!=\![\boldsymbol{S}_{1}\,|\,\boldsymbol{S}_{2}]\!=
\![\boldsymbol{I}\,|\,\boldsymbol{Q}]
\mleft[
\begin{array}{@{\,}c@{\,}|@{\,}c@{\,}}
\boldsymbol{C}_1 & \boldsymbol{0} \\
\hline
\boldsymbol{0}&\boldsymbol{C}_2
\end{array}
\mright]\!\! + \![\boldsymbol{0}\,|\,\boldsymbol{t}]\!\mleft[
\begin{array}{@{\,}c|c@{\,}}
\boldsymbol{1}_{N_1}^{\intercal} & \boldsymbol{0}_{N_2}^{\intercal} \\[0.5ex]
\hline
\boldsymbol{0}_{N_1}^{\intercal}&\boldsymbol{1}_{N_2}^{\intercal}
\end{array}
\mright]\!,\!\!\!\!\vspace{-1ex}
\end{equation}
where we have simplified the notation by omitting subscripts that can be inferred from context, which includes relabeling $\boldsymbol{Q} = \boldsymbol{Q}_{2}$ and $\boldsymbol{t} = \boldsymbol{t}_{2}$.

The problem addressed in this article is therefore to estimate, with basis on equations \eqref{eq:meas_full} and \eqref{eq:big_S_ego}, the rotation matrix $\boldsymbol{Q}$ and translation vector $\boldsymbol{t}$, given perfect knowledge of the conformation matrix $\boldsymbol{C}_1$ -- which implies exact knowledge of $\bm{D}_1$ -- and possession of an estimate of the matrix $\bm{D}_{12}$ subject to noise, under the egoistic condition that $\boldsymbol{C}_2$ is unknown and for a general case where $N_1 \neq N_2$.

    
\vspace{-1ex}
\subsection{A Note on Related SotA}
\label{sec:Q_est}
\vspace{-1ex}

To the best of our knowledge, the egoistic and generalized variation of the \ac{RBL} problem described above is original, but a related problem was considered in \cite{PizzoICASSP2016}, however with the assumptions that $N_1 = N_2$ and $\boldsymbol{C}_2$ is also known.
Unfortunately, a critical error\footnote{For the sake of completeness, a proof of incorrectness of the material in \cite[Subsec. 3.2]{PizzoICASSP2016} can be found in the journal version of this work \cite{Nic_Ego_journal}.} was made in \cite[Subsec. 3.2]{PizzoICASSP2016}, which makes the approach thereby ineffective for the estimation of the translation vector $\boldsymbol{t}$.
In spite of the aforementioned error, the method in \cite{PizzoICASSP2016} partially inspired the contribution of our article to be introduced subsequently, such that it is useful to briefly revise in the sequel the portion of the method regarding the estimation of the rotation matrix $\boldsymbol{Q}$.

First, consider the $N\times N$ classic Sch{\"o}nberg \ac{DCM}, defined by \cite{Torgerson_1952} 
\vspace{-0.5ex}
\begin{equation}
\label{eq:J}
\boldsymbol{J} = \boldsymbol{I}-\frac{1}{N}\boldsymbol{1}\boldsymbol{1}^\intercal.
\vspace{-0.5ex}
\end{equation}

Left- and right-multiplying a measured distance matrix by the \ac{DCM} $\boldsymbol{J}$, and scaling the result by $-\frac{1}{2}$, yields
\begin{equation}
\bar{\bm{D}}_{12}^{\odot 2}= -\frac{1}{2}\boldsymbol{J}\bm{D}_{12}^{\odot 2}\boldsymbol{J}=\boldsymbol{J}\boldsymbol{S}_1^\intercal\boldsymbol{S}_2\boldsymbol{J}=\boldsymbol{C}_{1}^\intercal\boldsymbol{Q}\boldsymbol{C}_{2}.
\label{eq:D_tilde}
\end{equation}

In order to facilitate the formulation of a problem to estimate $\boldsymbol{Q}$, it proves convenient to apply an \ac{OPP} onto equation \eqref{eq:D_tilde}, which under the assumption of perfect knowledge of $\boldsymbol{C}_{2}$ can be achieved by defining \cite{OPP_1966}
\begin{subequations}
\label{eq:right_mult}
\begin{equation}
\check{\bm{D}}_{12}^{\odot 2}\triangleq\bar{\bm{D}}_{12}^{\odot 2}\boldsymbol{C}_{2 }^\dag=\boldsymbol{C}_{1 }^\intercal\boldsymbol{Q},
\end{equation}
where
\begin{equation}
\boldsymbol{C}_{2}^\dag \triangleq \boldsymbol{C}_{2}^\intercal(\boldsymbol{C}_{2}\boldsymbol{C}_{2}^\intercal)^{-1}.
\end{equation}
\end{subequations}

Then, the relative rotation $\boldsymbol{Q}$ of body 2 with respect to the orientation of body 1 can be estimated by solving the problem
\begin{equation}
\hat{\boldsymbol{Q}}_{OPP}=\argmin_{\boldsymbol{Q}\in\mathbb{R}^{3\times 3}} ||\check{\bm{D}}_{12}^{\odot 2}-\boldsymbol{C}_{1}^\intercal\boldsymbol{Q}||_F^2,
\label{eq:OPP_opt}
\end{equation}
which can be obtained in closed form via \ac{SVD} of the matrix $\boldsymbol{C}_{1 }\check{\bm{D}}_{12}^{\odot 2}$.

In particular, the solution of problem \eqref{eq:OPP_opt} is given by \cite{PizzoICASSP2016}
\begin{subequations}
\label{eq:OPPsolution}
\begin{equation}
\hat{\boldsymbol{Q}}_{OPP}=\boldsymbol{U}\boldsymbol{V}^\intercal,
\end{equation}
with $\boldsymbol{U}$ and $\boldsymbol{V}$ such that
\begin{equation}
\boldsymbol{C}_{1 }\check{\bm{D}}_{12}^{\odot 2} = \boldsymbol{U}\boldsymbol{\Sigma}\boldsymbol{V}^\intercal.
\end{equation}
\end{subequations}

We emphasize that although it was assumed in \cite{PizzoICASSP2016} that both rigid bodies have the same number of landmark points ($e.g., N_1 = N_2$), the notion of a relative rotation \eqref{eq:big_S_ego} between two bodies of different shapes and number of landmark points is geometrically well defined, as can be inferred from equation \eqref{eq:big_S_ego}.
In particular, by aligning the rotation matrix of the first rigid body with the cartesian coordinates, such that $\bm{Q}_1 = \bm{I}$, the orientation $\bm{Q}_2$ of the second body with respect to the first, becomes simply the relative rotation itself.
In other words, $\bm{Q}_1 = \bm{I} \Longrightarrow \bm{Q}_2 = \bm{Q}$, or more generally, $\bm{Q} = \bm{Q}_1^\intercal\cdot\bm{Q}_2$.

\section{Proposed method}
\label{sec:prop}

The assumption of pre-existing knowledge of the conformation matrix $\bm{C}_2$, which is typical the \ac{SotA} \ac{RBL} methods \cite{PizzoICASSP2016, Chen_2015} is hard to meet in practical conditions.
In \ac{AD}-related \ac{V2X} applications, for instance, that would require that a vehicle attempting to locate other vehicles in its vicinity is aware of their shapes, an obviously impractical requirement given the enormous diversity in vehicle models, which are also constantly updated.
In order to mitigate this problem, we propose in this section methods to estimate $\bm{t}$ and $\bm{Q}$, respectively, without the requirement that $\bm{C}_2$ is known.

\subsection{Translation Estimation}
\label{sec:center_opt}

Let us start by pointing out that not knowing $\bm{C}_2$ implicates not knowing the intra-distances matrix $\bm{D}_2$.
And while the reverse implication is not logically true -- $i.e.$, in principle one could have knowledge of $\bm{D}_2$ but nor $\bm{C}_2$ -- the assumption that
$\bm{D}_2$ is also not available to the rigid body 1 is consistent with egoistic principle followed in this article, as indeed, an assumption of knowledge of $\bm{D}_2$ would require that the target vehicle broadcasts such information\footnote{Notice that an $N$-point \ac{3D} conformation matrix contains $3N$ entries, while the corresponding intra-distance matrix contains $N(N-1)/2$ distinct entries, such that the intra-distances data is larger than the conformation data for $N > 7$, which is a small number of points to define a rigid body in \ac{3D}.}.
In what follows, we therefore assume no knowledge of $\bm{D}_2$.

Under such conditions, the first problem at hand is one of matrix completion, and although several methods to solve such a problem exist \cite{Fang2012,Nguyen2019,Fan2024}, a number of which could be used, we here consider the simple and well-known Nystr\"om approximation method \cite{Williams_2001}, which applied to the \ac{EDM} $\bm{D}$ from equation \eqref{eq:full_D} yields\footnote{Note that the Nystr\"om approximation in general only works if the $\text{rank}(\bm{D}_1)\geq \text{rank}(\bm{D}_2)$, which means that the first body must have at least the same amount of sensors as the second body. If that condition is not satisfied, alternative matrix completion methods, $e.g.$ \cite{Fang2012,Nguyen2019,Fan2024}, may yield better results.} the following estimate of $\bm{D}_2$
\begin{equation}
\hat{\bm{D}}_2 \approx \mathbb{H}\big[\bm{D}_{12}^\intercal\bm{D}_1^{-1}\bm{D}_{12}\big],
\label{eq:NyAp}
\end{equation}
where $\mathbb{H}\big[\cdot\big]$ denotes a hollowing operator that enforces all elements of the diagonal matrix to be zero.

With possession of the intra-distances matrix of the first body $\bm{D}_1$, the measurements $\tilde{\bm{D}}_{12}$ corresponding the distances between the two bodies, and the latter estimate $\hat{\bm{D}}_2$ of the intra-distances matrix corresponding to the second rigid body, the full sample \ac{EDM} corresponding to all distances within and between the two rigid bodies can be reconstructed as
\begin{equation}
\hat{\bm{D}}=\mleft[
\begin{array}{c|c}
\bm{D}_1 & \tilde{\bm{D}}_{12} \\
\hline\\[-2ex]
\tilde{\bm{D}}_{12}^{\intercal}&\mathbb{H}\big[\tilde{\bm{D}}_{12}^\intercal\bm{D}_1^{-1}\tilde{\bm{D}}_{12}\big]
\end{array}
\mright],
\label{eq:full_D_est}
\end{equation}
such than an \ac{MDS}-based first estimate of the position of all sensors from both rigid bodies can be obtained as \cite{Torgerson_1952}
\begin{equation}
\label{eq:S_MDS}
[\hat{\bm{S}}^{*}_{1}, \hat{\bm{S}}^{*}_{2}] =\bm{V} \bm{\Lambda}^{1/2},
\end{equation}
where $\bm{V}$ and $\bm{\Lambda}$ are the eigenvector and eigenvalue pairs of the corresponding double-centered \ac{EDM}, that is
\begin{equation}
\bar{\bm{D}} = \bm{V} \bm{\Lambda} \bm{V}^\intercal,
\end{equation}
with
\begin{equation}
\bar{\bm{D}}= -\frac{1}{2}\boldsymbol{J}_{N_1+N_2}\hat{\bm{D}}^{\odot 2}\boldsymbol{J}_{N_1+N_2},
\end{equation}
where $\boldsymbol{J}_{N_1+N_2}$ is a $(N_1+N_2)$-point Sch{\"o}nberg \ac{DCM} build as per equation \eqref{eq:J}.

The initial \ac{MDS} solution given by equation \eqref{eq:S_MDS} can then be brought to the reference frame of the first rigid body via a Procrustes transformation
by solving
\begin{equation}
(\bm{Q}^{*},\bm{t}^{*}) = \hspace{-3ex}\argmin_{\bm{Q}\in\mathbb{R}^{3\times 3},\bm{t}\in\mathbb{R}^{3\times 1}} || \bm{C}_{1} - (\bm{Q}\,\hat{\bm{S}}^{*}_{1} + \bm{t}\otimes \bm{1}_{N_1}^\intercal) ||_F,
\end{equation}
from which we then obtain
\begin{equation}
\label{eq:S_est}
\hat{\bm{S}} = [\bm{C}_{1}, \hat{\bm{S}}_{2}] = [\bm{C}_{1}, \bm{Q}^{*}\hat{\bm{S}}^{*}_{2} + \bm{t}^{*}\otimes \bm{1}_{N_2}^\intercal],
\end{equation}
or, more explicitly
\begin{equation}
\label{eq:S2_est}
\hat{\bm{S}}_2 = \bm{Q}^{*}\hat{\bm{S}}^{*}_{2} + \bm{t}^{*}\otimes \bm{1}_{N_2}^\intercal.
\end{equation}

Substituting the latter result into equation \eqref{eq:big_S_ego} and using the relation $\boldsymbol{Q}_{2}\boldsymbol{C}_{2}=\hat{\boldsymbol{S}}_{2}\boldsymbol{J}_{N_2}$, we obtain
\begin{equation}
\label{eq:S_est_obj}
\begin{split}
\hat{\bm{S}}=\mleft[
\begin{array}{@{\,}c@{\,}|@{\,}c@{\,}}
\boldsymbol{C}_1 & \boldsymbol{0}_{3\times N_2}^{\intercal} \\
\hline
\boldsymbol{0}_{3\times N_1}^{\intercal}&(\bm{Q}^{*}\hat{\bm{S}}^{*}_{2}\! +\! \bm{t}^{*}\!\otimes\!\bm{1}_{N_2}^\intercal)\boldsymbol{J}_{N_2}
\end{array}
\mright]\! +\! [\boldsymbol{0}|\boldsymbol{t}]\!\mleft[
\begin{array}{@{\,}c|c@{\,}}
\boldsymbol{1}_{N_1}^{\intercal} & \boldsymbol{0}_{N_2}^{\intercal} \\[0.5ex]
\hline
\boldsymbol{0}_{N_1}^{\intercal}&\boldsymbol{1}_{N_2}^{\intercal}
\end{array}
\mright]\!,
\end{split}
\end{equation}

Utilizing the latter expression, we can finally formulate a quadratic program to find the translation vector $\boldsymbol{t}$, namely
\begin{equation}
\hat{\boldsymbol{t}}=\argmin_{\boldsymbol{t}}||\boldsymbol{J}_{N_1+N_2}(\hat{\bm{S}}^\intercal\hat{\bm{S}}+\tfrac{1}{2}\hat{\bm{D}}^{\odot 2})\boldsymbol{J}_{N_1+N_2}||^2_F,
\label{eq:center_opt}
\end{equation}
which can easily be solved by common optimization tools, such as gradient descent or interior point methods \cite{Nocedal1999,Ruder2016}.
    
\subsection{Rotation Matrix Estimation}
\label{sec:Q_C_est}

With the estimate $\hat{\bm{S}}_2$ obtained via equation \eqref{eq:S2_est} in hands, a robust estimate of the rotation matrix $\bm{Q}$ corresponding to the second rigid body can be obtained via a procedure similar to that described in Subsection \ref{sec:Q_est}.

Before we proceed, let us emphasize that, in principle, $\hat{\bm{Q}}$ can be extracted by
\begin{equation}
\label{eq:Q_est_eig}
(\hat{\bm{S}}_2\bm{J}_{N_2})(\hat{\bm{S}}_2\bm{J}_{N_2})^\intercal = \bm{Q} \bm{\Lambda} \bm{Q}^\intercal
= \bm{Q} \boldsymbol{C}_{2}\boldsymbol{C}_{2}^\intercal \bm{Q}^\intercal,
\end{equation}
where we used $\boldsymbol{Q}_{2}\boldsymbol{C}_{2}=\hat{\boldsymbol{S}}_{2}\boldsymbol{J}_{N_2}$ in the last equality.

Notice, however, that the eigenvalue decomposition in equation \eqref{eq:Q_est_eig} is such that the eigenvectors are ordered according to their corresponding eigenvalues, which in turn relate to the largest orthogonal dimensions of the body \cite{jolliffe2002principal,hastie2009elements}.
It follows that the columns of the estimate obtained via equation \eqref{eq:Q_est_eig} may be swapped for rigid bodies with approximately spherical shapes, leading to large estimation errors.

We therefore propose instead the following method.
First, let us return to equation \eqref{eq:D_tilde}, but this time accounting for the fact that $\bm{S}_1$ and $\bm{S}_2$ have different numbers $N_1$ and $N_2$ of landmark points, such that
\vspace{-0.5ex}
\begin{equation}
\bar{\bm{D}}_{12}^{\odot 2}= -\frac{1}{2}\boldsymbol{J}_{N_1}\bm{D}_{12}^{\odot 2}\boldsymbol{J}_{N_2}=\boldsymbol{C}_{1}^\intercal\boldsymbol{Q}\boldsymbol{C}_{2},
\vspace{-0.5ex}
\end{equation}
which if left-multiply by the pseudo-inverse of $\boldsymbol{C}_{1}^\intercal$ yields
\vspace{-0.5ex}
\begin{subequations}
\label{eq:left_mult}
\begin{equation}
\check{\bm{D}}_{12}^{\odot 2}\triangleq\boldsymbol{C}_{1}^\dag\bar{\bm{D}}_{12}^{\odot 2}= \bm{Q}\boldsymbol{C}_{2},
\vspace{-0.5ex}
\end{equation}
where
\vspace{-0.5ex}
\begin{equation}
\boldsymbol{C}_{1}^\dag \triangleq (\boldsymbol{C}_{1}\boldsymbol{C}_{1}^\intercal)^{-1}\boldsymbol{C}_{1}.
\vspace{-0.5ex}
\end{equation}
\end{subequations}

Then, squaring equation \eqref{eq:left_mult} yields
\vspace{-0.5ex}
\begin{equation}
\check{\bm{D}}_{12}^{\odot 2}\check{\bm{D}}_{12}^{\odot 2 \intercal}= \bm{Q}\boldsymbol{C}_{2}\boldsymbol{C}_{2}^\intercal\bm{Q}^\intercal = \bm{Q}\boldsymbol{\Lambda}\bm{Q}^\intercal,
\vspace{-0.5ex}
\end{equation}
from which the following optimization problem can be constructed
\vspace{-0.5ex}
\begin{equation}
\hat{\boldsymbol{Q}}=\argmin_{\boldsymbol{Q}} ||\check{\bm{D}}_{12}^{\odot 2}\check{\bm{D}}_{12}^{\odot 2 \intercal}-\bm{Q}\bm{\Lambda}\bm{Q}^\intercal||_F^2.
\label{eq:OPP_opt_2}
\vspace{-0.5ex}
\end{equation}

We emphasize that although the solution of problem \eqref{eq:OPP_opt_2} can be easily obtained via common optimization theory tools \cite{Nocedal1999,Ruder2016}, the result can also be severely degraded by the order of the eigenvalues in $\bm{\Lambda}$.
Fortunately, however, in \ac{3D} there are only 6 distinct permutations of $\bm{\Lambda}$, such that the solution with the permutation that yields the smallest objective can be estimated as the correct one.


\section{Performance Evaluation}
\label{sec:res}

\begin{figure}[H]
\centering
\includegraphics[width=\columnwidth]{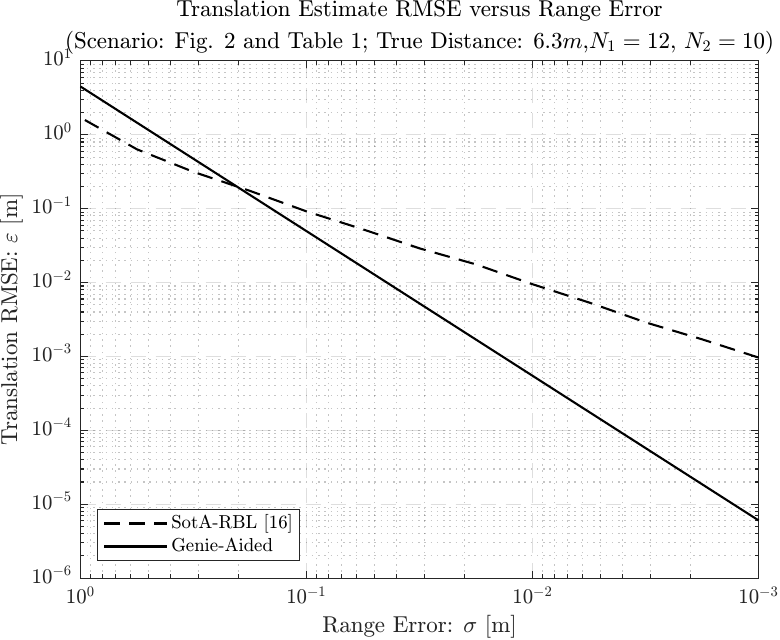}
\vspace{-4ex}
\caption{\ac{RMSE} of the translation estimate of the \acf{GA} proposed method and the \ac{SotA}, over the range error $\sigma$.}
\label{fig:TraRes_Genie}
\end{figure}

In this section we provide simulation results illustrating the performance of the contributed egoistic \ac{MDS}-based \ac{RBL} technique.
Since, to the best of our knowledge, no equivalent \ac{SotA} method exists for the egoistic set-up here considered in which $\bm{C}_2$ is unknown, we first compare in Figure \ref{fig:TraRes_Genie} only results on translation vector estimation via the non-egoistic method of \cite{Chen_2015}, against proposed technique of Subsection \ref{sec:center_opt}, but using the estimate of $\bm{Q}$ from \cite{Chen_2015} in equation \eqref{eq:S_est_obj}.
For the sake of disambiguation, the corresponding results of the proposed method with an externally fed rotation matrix is referred to as the ``Genie-Aided'' scheme.

\begin{table*}[ht]
\centering
\caption{Simulation Parameters}
\vspace{-2ex}
\begin{tabular}[H]{|c|c|}
\hline
\Tstrut
\begin{tabular}{@{}c@{}c@{}}
Reference frames
\end{tabular} &
\begin{tabular}{@{}c@{}c@{}}

\setlength{\arraycolsep}{2pt} 
$\boldsymbol{C}_1=\mleft[
\begin{array}{*{12}{c}}
-1.25 & 1.25 & -1.25 & 1.25 & -1.25 & 1.25 & -1.25 & 1.25 & -1.25 & 1.25 & -1.25 & 1.25 \\
-4 & -4 & -4 & -4 & 0 & 0 & 0 & 0 & 4 & 4 & 4 & 4 \\
0.5 & 0.5 & 1 & 1 & 1 & 1 & 4 & 4 & 4 & 4 & 0.5 & 0.5 \\
\end{array}
\mright] $\\

\setlength{\arraycolsep}{2.5pt} 
$\boldsymbol{C}_2=\mleft[
\begin{array}{*{12}{c}}
-1 & 1 & -1 & 1 & -1 & 1 & -1 & 1 & -1 & 1 \\
2 & 2 & 1 & 1 & -1 & -1 & -2 & -2 & 0 &0  \\
1 & 1 & 1.5 & 1.5 & 1.5 & 1.5 & 1 & 1 & 0.5 & 0.5 \\
\end{array}
\mright] $

\end{tabular}
\Tstrut\\[2ex]
\hline
\Tstrut
\begin{tabular}{@{}c@{}}
Translations
\end{tabular} &
\begin{tabular}{@{}c@{}c@{}}
$\boldsymbol{t}_1=[0, 0, 0]^\intercal$\Tstrut\\
$\boldsymbol{t}_2=\boldsymbol{t}=[7,3, 0.5]^\intercal$\Tstrut
\vspace{1ex}
\end{tabular}\Tstrut\\
\hline
Rotations  &
\begin{tabular}{@{}c@{}c@{}}
$[\psi_1,\theta_1,\phi_1]=[0^\circ,0^\circ,0^\circ]$ \Tstrut\\
$[\psi_2,\theta_2,\phi_2]=[10^\circ,20^\circ,45^\circ]$
\end{tabular}\\
\hline
\end{tabular}
\label{table::parameters}
\vspace{-2ex}
\end{table*}

The performance metric of choice is the \ac{RMSE}, as a function of the ranging error\footnote{Note that the ranging error is not equivalent to the exact error in meters but rather the error used in the noise calculations given in \eqref{eq:sqr_meas}.} $\sigma$, namely
\begin{equation}
\varepsilon = \sqrt{\frac{1}{K}\sum_{k=1}^{K}|\hat{\boldsymbol{t}}^{(k)}-{\boldsymbol{t}}|_2^2},
\end{equation}
where $\hat{\boldsymbol{t}}^{(k)}$ denotes an estimate obtained at a $k$-th realization, and we emphasize that the dependence of $\varepsilon$ on $\sigma$ is due to the errors $\hat{\boldsymbol{t}}^{(k)}$, not included explicit in the notation for simplicity.

\vspace{-4ex}
\begin{figure}[H]
\centering
\includegraphics[width=\columnwidth]{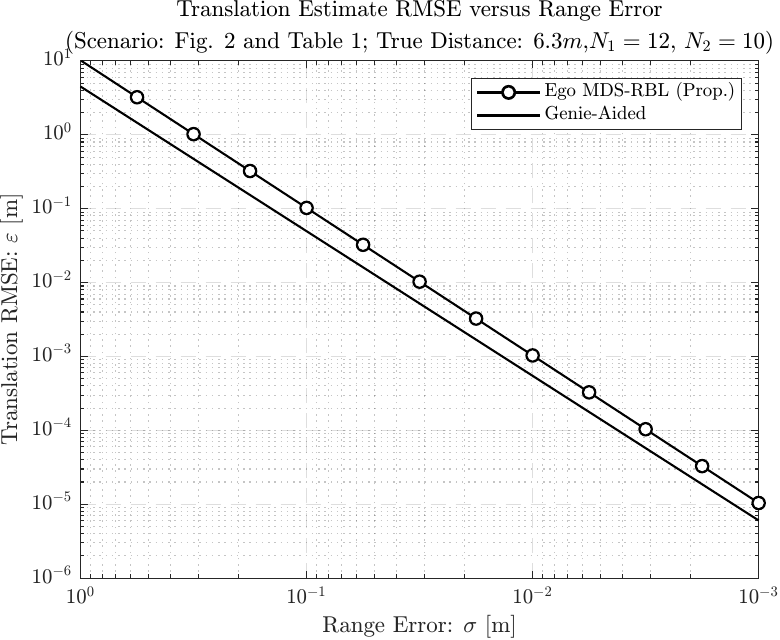}
\vspace{-4ex}
\caption{\ac{RMSE} of the translation estimate of the proposed method compared to the GA variation, over the range error $\sigma$.}
\label{fig:TraRes_Ego}
\end{figure}

Each point in the figure is obtained by averaging $K=10^3$ Monte-Carlo realizations, using the system parameters described in Table \ref{table::parameters}.
The algorithms are implemented in MATLAB, with the minimization problems solved using the CVX optimization package.

The results in Figure \ref{fig:TraRes_Genie} show that in a non-egoistic scenario, the proposed method outperforms the \ac{SotA} alternative if ranging errors are below $20$ cm, which is well within the typical values of sensing technology used in the Automotive Industry \cite{MalekianSenJ2018}.

Finally, a comparison between the latter Genie-Aided method and the actually proposed egoistic scheme is offered in Figure \ref{fig:TraRes_Ego}, which confirms that the contributed (egoistic) method maintains a performance close to that of the Genie-Aided (non-egoistic) alternative.

\vspace{-4ex}
\section{Conclusion}
%
%
%
%
%

We proposed a novel anchorless \ac{RBL} algorithm suitable for application in \ac{AD}, which enables a rigid body to egoistically detect the relative translation (effective distance) and orientation (relative rotation) of another body, based only on a set of measurements of the distances between sensors of one vehicle to the other and without knowledge of the shape of the latter.
A key point of the proposed method is that the translation vector between the two-bodies is modeled using the \ac{MDS} double-centering operator, enabling its applicability between rigid bodies of different shapes, in contrast to conventional approaches which require both bodies to have the same shape.
Simulation results illustrate the good performance of the proposed technique in terms of \ac{RMSE} as a function of the ranging error, in the desired (and typical) moderate to low ranging errors regime.

\vspace{-2ex}


\begin{thebibliography}{10}
    \providecommand{\url}[1]{#1}
    \csname url@samestyle\endcsname
    \providecommand{\newblock}{\relax}
    \providecommand{\bibinfo}[2]{#2}
    \providecommand{\BIBentrySTDinterwordspacing}{\spaceskip=0pt\relax}
    \providecommand{\BIBentryALTinterwordstretchfactor}{4}
    \providecommand{\BIBentryALTinterwordspacing}{\spaceskip=\fontdimen2\font plus
    \BIBentryALTinterwordstretchfactor\fontdimen3\font minus
      \fontdimen4\font\relax}
    \providecommand{\BIBforeignlanguage}[2]{{%
    \expandafter\ifx\csname l@#1\endcsname\relax
    \typeout{** WARNING: IEEEtran.bst: No hyphenation pattern has been}%
    \typeout{** loaded for the language `#1'. Using the pattern for}%
    \typeout{** the default language instead.}%
    \else
    \language=\csname l@#1\endcsname
    \fi
    #2}}
    \providecommand{\BIBdecl}{\relax}
    \BIBdecl
    
    \bibitem{Yassin_2016}
    A.~Yassin \emph{et. al}, ``Recent advances in indoor localization: A survey on
      theoretical approaches and applications,'' \emph{IEEE Communications Surveys
      \& Tutorials}, vol.~19, no.~2, 2017.
    
    \bibitem{VoCST2016}
    Q.~D. Vo and P.~De, ``A survey of fingerprint-based outdoor localization,''
      \emph{IEEE Communications Surveys \& Tutorials}, vol.~18, no.~1, 2016.
    
    \bibitem{Nic:RSSI}
    N.~Führling \emph{et. al}, ``Robust
      received signal strength indicator {(RSSI)}-based multitarget localization
      via gaussian process regression,'' \emph{IEEE J. Indoor Seamless Position. Navig.}, vol.~1, 2023.
    
    \bibitem{Al-SadoonTAP2020}
    M. Al-Sadoon \emph{et. al}, ``{AOA} localization for vehicle-tracking systems using a dual-band
      sensor array,'' \emph{IEEE Trans. Antennas Propag.},
      vol.~68, no.~8, 2020.
    
    \bibitem{ZengTSP2022}
    G.~Zeng \emph{et. al}, ``Global and asymptotically
      efficient localization from range measurements,'' \emph{IEEE Trans. on
      Signal Processing}, vol.~70, 2022.
    
    \bibitem{burghal_2020}
    D.~Burghal \emph{et. al}, ``A
      comprehensive survey of machine learning based localization with wireless
      signals,'' 2020.
    
    \bibitem{Zhang_2021}
    J.~A. Zhang \emph{et. al}, ``An overview of signal processing techniques for joint
      communication and radar sensing,'' \emph{IEEE J. Sel. Topics Signal Process.}, vol.~15, no.~6, 2021.
    
    \bibitem{Rayan_2024}
    K.~R.~R. Ranasinghe \emph{et. al}, ``Fast and efficient
      sequential radar parameter estimation in {MIMO-OTFS} systems,'' in
      \emph{IEEE International Conference on Acoustics, Speech
      and Signal Processing (ICASSP)}, 2024.
    
    \bibitem{Rayan_Journal}
    K.~R.~R. Ranasinghe \emph{et. al},
      ``Joint channel, data and radar parameter estimation for {AFDM} systems in
      doubly-dispersive channels,'' 2024.
    
    \bibitem{WangTSP2020}
    Y.~Wang \emph{et. al}, ``An investigation and
      solution of angle based rigid body localization,'' \emph{IEEE Trans. on
      Signal Processing}, vol.~68, 2020.
    

    \bibitem{FuehrlingV2X2024}
    \BIBentryALTinterwordspacing
    N.~F{\"u}hrling \emph{et. al},
      ``Enabling Next-Generation {V2X} Perception: Wireless Rigid Body Localization
      and Tracking,'' \emph{arXiv preprint arXiv:2408.00349}, 2024.
    \BIBentrySTDinterwordspacing
    
    \bibitem{AHMED_2020}
    F.~Ahmed \emph{et. al}, ``Comparative study of
      seamless asset location and tracking technologies,'' \emph{Procedia
      Manuf.}, International Conference
      on Flexible Automation and Intelligent Manufacturing, vol.~51, 2020.
    
    \bibitem{Bruk_2023}
    B.~Gebregziabher, ``Multi object tracking for predictive collision avoidance,''
      2023.
    
    \bibitem{eckenhoff_2019}
    K.~Eckenhoff, Y.~Yang, P.~Geneva, and G.~Huang, ``Tightly-coupled
      visual-inertial localization and {3-D} rigid-body target tracking,''
      \emph{IEEE Robotics and Automation Letters}, vol.~4, no.~2,
      2019.
    
    \bibitem{Huang_2022}
    Y.~Huang, J.~Du, Z.~Yang, Z.~Zhou, L.~Zhang, and H.~Chen, ``A survey on
      trajectory-prediction methods for autonomous driving,'' \emph{IEEE
      Trans. on Intelligent Vehicles}, vol.~7, no.~3, 2022.
    
    \bibitem{Chen_2015}
    S.~Chen and K.~C. Ho, ``Accurate localization of a rigid body using multiple
      sensors and landmarks,'' \emph{IEEE Trans. on Signal Processing},
      vol.~63, no.~24, 2015.
    
    \bibitem{Huang_2019}
    J.~Huang \emph{et. al}, ``{ClusterSLAM}: A {SLAM}
      backend for simultaneous rigid body clustering and motion estimation,'' in
      \emph{IEEE/CVF International Conference on Computer Vision (ICCV)},
      2019.
    
    \bibitem{Barros_2022}
    A.~Macario~Barros \emph{et. al}, ``A
      comprehensive survey of visual {SLAM} algorithms,'' \emph{Robotics}, vol.~11,
      no.~1, 2022.
    
    \bibitem{Bavle_2023}
    H.~Bavle \emph{et. al}, ``From
      {SLAM} to situational awareness: Challenges and survey,'' \emph{Sensors},
      vol.~23, no.~10, 2023.
    
    \bibitem{Seref_2013}
    S.~Sagiroglu and D.~Sinanc, ``Big data: A review,'' in \emph{International
      Conference on Collaboration Technologies and Systems (CTS)}, 2013.
    
    \bibitem{Bras_2016}
    S.~Brás \emph{et. al}, ``Nonlinear
      observer for {3D} rigid body motion estimation using doppler measurements,''
      \emph{IEEE Trans. on Automatic Control}, vol.~61, no.~11, 2016.
    
    \bibitem{PizzoICASSP2016}
    A.~Pizzo, S.~P. Chepuri, and G.~Leus, ``Towards multi-rigid body
      localization,'' in \emph{IEEE International Conference on Acoustics,
      Speech and Signal Processing (ICASSP)}, 2016.
    
    \bibitem{Chepuri_2013}
    S.~P. Chepuri \emph{et. al}, ``Tracking
      position and orientation of a mobile rigid body,'' in \emph{5th IEEE
      International Workshop on Computational Advances in Multi-Sensor Adaptive
      Processing (CAMSAP)}, 2013.

      
      \bibitem{Nic_Ego_journal}
    \BIBentryALTinterwordspacing
    N.~F{\"u}hrling \emph{et. al},
      ``Robust Egoistic Rigid Body Localization,'' \emph{arXiv preprint arXiv:2501.10219}, 2025.
    \BIBentrySTDinterwordspacing
    
    \bibitem{Torgerson_1952}
    W.~S. Torgerson, ``Multidimensional scaling: I. theory and method,''
      \emph{Psychometrika}, vol.~17, no.~4, Dec. 1952.
    
    \bibitem{OPP_1966}
    P.~Schönemann, ``A generalized solution of the orthogonal procrustes
      problem,'' \emph{Psychometrika}, vol.~31, no.~1, 1966.
    
    \bibitem{Fang2012}
    H.~Fang and D.~P. O'Leary, ``Euclidean distance matrix completion problems,''
      \emph{Optimization Methods and Software}, vol.~27, no. 4-5,
      2012.
    
    \bibitem{Nguyen2019}
    L.~T. Nguyen, J.~Kim, and B.~Shim, ``Low-rank matrix completion: A contemporary
      survey,'' \emph{IEEE Access}, vol.~7, 2019.
    
    \bibitem{Fan2024}
    Y.~Fan and M.~Pesavento, ``Localization in sensor networks using distributed
      low-rank matrix completion,'' in \emph{IEEE International
      Conference on Acoustics, Speech and Signal Processing (ICASSP)}, 2024.
    
    \bibitem{Williams_2001}
    C.~K.~I. Williams and M.~Seeger, ``Using the {Nystr\"{o}m} method to speed up
      kernel machines,'' in \emph{Proc. of the 13th International Conference
      on Neural Information Processing Systems}, ser. NIPS'00. Cambridge, MA, USA: MIT Press, 2000.
    
    \bibitem{Nocedal1999}
    J.~Nocedal and S.~J. Wright, \emph{Numerical Optimization}, ser. Springer
      Series in Operations Research and Financial Engineering.\hskip 1em plus 0.5em
      minus 0.4em\relax Springer New York, NY, 1999.
    
    \bibitem{Ruder2016}
    S.~Ruder, ``An overview of gradient descent optimization algorithms,''
      \emph{arXiv preprint arXiv:1609.04747}, 2016.
    
    \bibitem{jolliffe2002principal}
    I.~T. Jolliffe, \emph{Principal component analysis for special types of
      data}.\hskip 1em plus 0.5em minus 0.4em\relax Springer, 2002.
    
    \bibitem{hastie2009elements}
    T.~Hastie, R.~Tibshirani, J.~H. Friedman, and J.~H. Friedman, \emph{The
      elements of statistical learning: data mining, inference, and
      prediction}.\hskip 1em plus 0.5em minus 0.4em\relax Springer, 2009, vol.~2.
    
    \bibitem{MalekianSenJ2018}
    R.~Malekian, K.~Curran, C.~F. Pedersen, B.~Cao, and X.~Qi, ``Guest editorial
      special issue on sensor technologies for connected cars: Devices, systems and
      modeling,'' \emph{IEEE Sensors Journal}, vol.~18, no.~12,
      2018.
    
    \end{thebibliography}
\end{document}